%% file: main.tex
\documentclass{article}

\usepackage[preprint]{style/Styles/neurips_2025}

\usepackage[utf8]{inputenc}
\usepackage[T1]{fontenc}
\usepackage{hyperref}
\usepackage{url}
\usepackage{booktabs}
\usepackage{amsfonts}
\usepackage{nicefrac}
\usepackage{microtype}
\usepackage{xcolor}
\usepackage{graphicx}
\usepackage{amsmath}
\usepackage{amssymb}
\usepackage{placeins}
\usepackage{caption}
\usepackage{float}
\usepackage{array}

\hypersetup{hidelinks}

\makeatletter
\newcommand{\Needspace}[1]{%
  \par
  \ifdim\dimexpr\pagegoal-\pagetotal\relax<#1
    \newpage
  \fi
}
\makeatother

\title{Dense vs Sparse Pretraining at Tiny Scale: Active-Parameter vs Total-Parameter Matching}
\author{%
  Abdalrahman Wael \\
  Independent Researcher \\
  \texttt{abdalrahmann.wael@gmail.com}
}
\date{}

\begin{document}

\maketitle

\input{sections/abstract}
\input{sections/introduction}
\input{sections/related_work}
\input{sections/method}
\input{sections/results}
\input{sections/limitations}
\input{sections/conclusion}

\FloatBarrier
\clearpage
\appendix
\input{sections/appendix}

\clearpage
\FloatBarrier
\bibliographystyle{plainnat}
\bibliography{refs}

\end{document}

%% file: sections/abstract.tex
\begin{abstract}
We study dense and mixture-of-experts (MoE) transformers in a tiny-scale pretraining regime under a shared LLaMA-style decoder training recipe. The sparse model replaces dense feed-forward blocks with Mixtral-style routed experts. Dense baselines are modestly width-resized to tightly match either active or total parameter budgets, while tokenizer, data, optimizer, schedule, depth, context length, normalization style, and evaluation protocol are held fixed. Our best sparse recipe uses four experts, top-2 routing, Switch-style load balancing, and router z-loss. In a three-seed full-data comparison, the dense active-match model reaches $1.6545 \pm 0.0012$ best validation loss, the MoE reaches $1.5788 \pm 0.0020$, and the dense total-match model reaches $1.5608 \pm 0.0025$. This yields a matched-active gap of $0.0758 \pm 0.0021$ in the MoE's favor and a matched-total gap of $0.0180 \pm 0.0020$ in the dense model's favor. Across training, the matched-active advantage grows while the matched-total dense advantage narrows sharply. In this sub-25M-parameter regime, MoE therefore improves validation loss under active-parameter matching but does not surpass dense training at equal total stored capacity.
\end{abstract}

%% file: sections/introduction.tex
\section{Introduction}

Mixture-of-experts (MoE) transformers decouple total parameter count from per-token active capacity: a model can store a larger total capacity while activating only a subset of expert parameters on each token. At large scale, this is often presented as a route to better quality through conditional computation. However, much of the strongest evidence comes from distributed settings where routing behavior, communication cost, and systems design are tightly entangled, making it harder to isolate the modeling contribution itself.

This paper asks a simpler empirical question: in a tiny, single-GPU-accessible regime, does replacing dense feed-forward blocks with sparse experts produce a real modeling benefit under controlled budgets? We study this question in a LLaMA-style decoder stack trained on TinyStories. The sparse model replaces dense feed-forward networks with Mixtral-style routed experts. Dense baselines are modestly width-resized to tightly match active or total parameter budgets, while the tokenizer, data pipeline, optimizer, schedule, depth, context length, normalization style, and evaluation protocol are held fixed.

Dense-versus-sparse fairness is not unique. In the matched-active comparison, the dense baseline is resized so that its overall active per-token parameter count approximately matches the MoE's active routed path. In the matched-total comparison, the dense baseline is resized so that its total stored parameter count matches the full MoE.

Our central result is straightforward. In the matched-active comparison, the stabilized top-2 MoE clearly outperforms the dense baseline. In the matched-total comparison, the dense baseline remains slightly stronger, but the gap is small and narrows with training. The resulting picture is not that sparse training dominates dense training outright, but that conditional computation is already useful in this tiny-scale regime when fairness is defined against active per-token parameters rather than total stored parameters.

The paper also clarifies which ingredients matter for sparse training at this scale. Naive top-1 routing without balancing collapses almost immediately. Adding Switch-style balancing removes that failure mode, and moving from top-1 to top-2 routing produces the main sparse quality gain. Router z-loss mostly improves routing behavior and stability diagnostics rather than serving as the primary source of the final validation gain.

We treat this as a controlled empirical study rather than a new architecture paper. The main contribution is to separate matched-active and matched-total conclusions, show that they lead to different dense-versus-sparse rankings, and document that the sparse gain is accompanied by healthy routing diagnostics rather than collapsed or pathological expert usage. In a sub-25M-parameter TinyStories setting, this provides a compact case study of how dense-versus-sparse conclusions change when active-parameter and total-parameter matching are kept distinct.

%% file: sections/related_work.tex
\section{Related Work}

Modern sparse expert models build on the conditional-computation framing introduced by sparsely gated MoE layers~\cite{shazeer2017outrageously}, which emphasized both the potential of expert specialization and the risk that routing can become imbalanced or degenerate. Switch Transformers~\cite{fedus2022switch} simplified this line with top-1 routing and a lightweight auxiliary balancing objective, showing that sparse models could be trained at very large parameter counts with a relatively simple routing mechanism. ST-MoE~\cite{zoph2022stmoe} then focused more directly on sparse-training stability, especially the role of router regularization terms such as the router z-loss.

Mixtral~\cite{jiang2024mixtral} is the closest architectural reference point for our sparse intervention because the MoE model changes the feed-forward block inside an otherwise standard decoder block. We do not propose a new sparse architecture. Instead, we use a Mixtral-style intervention and compare it against width-resized dense baselines to ask when sparse conditional computation helps at tiny scale and how the answer depends on the fairness criterion.

For broader motivation, GLaM~\cite{du2022glam} and large-scale autoregressive MoE studies~\cite{artetxe2022efficient} support the view that MoEs can improve quality relative to per-token active work, while TinyStories~\cite{eldan2023tinystories} provides a deliberately small language-modeling regime suitable for controlled architectural comparisons. We use TinyStories not as a claim about broad downstream transfer, but as a way to isolate dense-versus-sparse pretraining behavior in a setting where small models can still exhibit meaningful language modeling performance.

Recent scaling-law work studies MoE behavior through separate parameter and compute axes. Clark et al.~\cite{clark2022unified} model routed language models using both parameter count and computational requirement, while later work studies MoE granularity, active-parameter count, total-parameter count, sparsity, and training compute at larger scales~\cite{krajewski2024scaling,abnar2025parameters,ludziejewski2025joint}. OLMoE~\cite{muennighoff2025olmoe} provides large-scale open evidence for dropless MoE training with load balancing, z-loss, and public training artifacts. Recent equal-resource studies also show that larger, tuned MoE regimes can outperform dense models under stricter total-parameter and compute constraints~\cite{li2026moe_equal_resource}. Our study is complementary: it does not derive a new scaling law or architecture, but gives a tiny-scale budget-matched comparison under a shared training recipe, with both matched-active and matched-total dense baselines reported.

We treat systems-oriented sparse execution work such as MegaBlocks~\cite{gale2023megablocks} as important secondary context rather than a main modeling contribution. Likewise, we use Chinchilla~\cite{hoffmann2022training} mainly as background for token-budget and undertraining considerations. The present paper is best read as an empirical fairness and training-dynamics study, not as a cluster-scale systems paper.

For contrast with later large-scale sparse-design lines, recent systems-aware MoE work such as DeepSeekMoE~\cite{dai2024deepseekmoe} and DeepSeek-V2~\cite{deepseekai2024deepseekv2} explores more aggressive expert specialization and efficiency tradeoffs at a very different operating scale from the controlled tiny-scale regime studied here.

%% file: sections/method.tex
\section{Method}

We compare dense and sparse models in a decoder-only LLaMA-style training stack with the same tokenizer, dataset, optimizer, schedule, depth, context length, normalization style, and evaluation protocol. The sparse intervention replaces dense FFNs with routed MoE feed-forward modules. The dense baselines are modestly resized for active- or total-parameter matching, so $d_{\mathrm{model}}$, query-head count, and attention parameter counts differ across model families.

\begin{figure}[!htbp]
\centering
\IfFileExists{figures/architecture_dense_vs_moe.pdf}{
  \includegraphics[width=\linewidth]{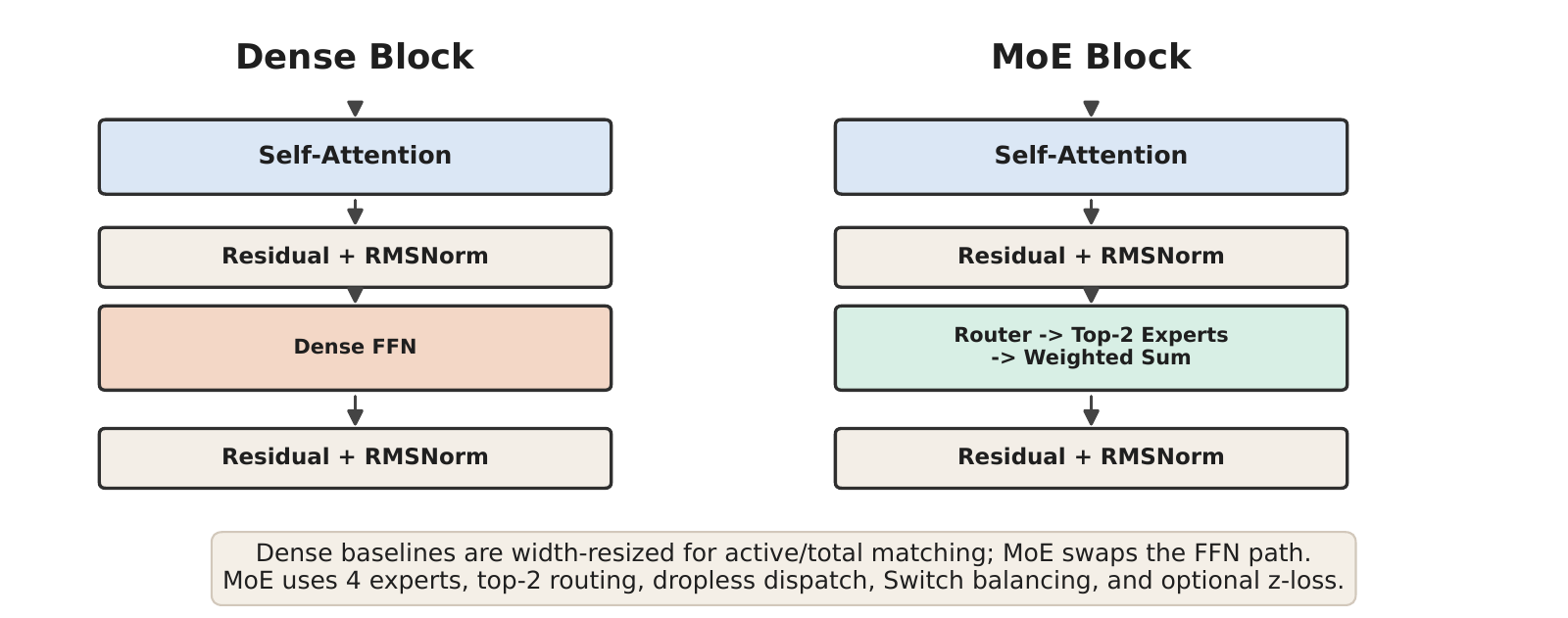}
}{
  \fbox{\parbox[c][1.8in][c]{0.95\linewidth}{\centering Placeholder for architecture figure\\same attention and residual path; only FFN replaced\\router $\rightarrow$ top-2 experts $\rightarrow$ weighted sum}}
}
\caption{Schematic comparison of a dense FFN block and a routed MoE FFN block. In the headline experiments, all models share the same training recipe and decoder-block style, while dense baselines are width-resized for active- or total-parameter matching.}
\label{fig:arch}
\end{figure}

Our main sparse recipe uses four experts, top-2 routing, dropless dispatch, Switch-style load balancing, and optional router z-loss stabilization. We do not use classic Switch-style token dropping or capacity truncation. This keeps the modeling comparison focused on sparse routing behavior rather than on execution-path shortcuts. In a separate appendix-side systems study, we evaluate more optimized grouped and stacked single-GPU dispatch paths, but those analyses are not the basis of the paper's central modeling claims.

The total training objective combines next-token cross-entropy with a load-balancing auxiliary term and, optionally, a router z-loss term:
\[
L_{\text{total}} = L_{\text{CE}} + \lambda_{\text{bal}} L_{\text{bal}} + \lambda_z L_z.
\]
Here $L_{\text{CE}}$ is the language-model cross-entropy, $L_{\text{bal}}$ encourages balanced expert usage and prevents immediate routing collapse, and $L_z$ penalizes overly large router logit scale. In the current best sparse recipe, $\lambda_{\text{bal}} = 10^{-2}$ and $\lambda_z = 10^{-3}$.

We evaluate under two fairness criteria. In the matched-active comparison, the dense baseline is resized so that its active parameter count approximately matches the MoE model's active routed path per token. In the matched-total comparison, the dense baseline is resized so that its total stored parameter count approximately matches the full MoE model. These comparisons answer different questions and should not be conflated.

Throughout the paper, active parameters means the shared non-expert parameters plus the dense FFN or top-2 routed expert path exercised per token; router weights are negligible at the displayed precision. We use active parameter count as the primary fairness proxy for per-token routed capacity; we do not claim exact FLOP matching. Under this definition, the three headline configurations are a dense active-match baseline with approximately 14.89M parameters, a top-2 MoE with approximately 21.06M total parameters and 14.77M active parameters, and a dense total-match baseline with approximately 21.06M parameters. The budget matching is tight at this tiny scale: the active-match counts differ by less than 1\%, and the total-match counts differ by less than 0.1\%.

Our primary metrics are held-out validation loss and perplexity. Routing diagnostics are used to show that the sparse gain is not explained by collapsed or pathological routing, while throughput and execution-layout analyses are treated as appendix-side support. Qualitative generations are included only as anecdotal checks and do not replace the main metric story.

Token budgets count tokens contributing to the next-token loss. With 512-token windows, each training row contributes 511 prediction targets after shifting the input and target streams by one position.

%% file: sections/results.tex
\section{Results}

The first result is that tiny-scale MoE does not work naively in this stack. A top-1 MoE without balancing collapses almost immediately, with most layers routing nearly all tokens to a single expert after only a short smoke run. Adding a Switch-style load-balancing auxiliary term removes this immediate failure mode and produces meaningfully spread expert fractions across layers. Once routing is stabilized, top-2 routing provides the main sparse quality gain over top-1. Router z-loss then acts primarily as a behavior-shaping refinement: it reduces router log-z and moderates later-layer sharpness, but it is not the main reason the sparse model works.

\begin{table}[!htbp]
\centering
\small
\caption{Diagnostic recipe-selection runs, not a fully budget-matched ablation sweep. Losses should only be compared directly within the same 40.5M-token setting; smoke runs are included to document routing failure and stabilization behavior. Validation loss is pure next-token cross-entropy.}
\label{tab:routing-ablation}
\resizebox{\linewidth}{!}{%
\begin{tabular}{lcccl}
\toprule
Variant & Budget & Balance & z-loss & Result \\
\midrule
Top-1 MoE & 100-step smoke & no & no & Collapsed routing; most layers routed nearly all tokens to one expert. \\
Top-1 MoE & 1.64M tokens & yes & no & Stable smoke run, val $5.3775$, max busiest expert fraction $0.45$. \\
Top-1 MoE & 40.5M tokens & yes & no & Stable longer run, final val $2.1981$. \\
Top-2 MoE & 40.5M tokens & yes & no & Final val $2.1618$; main sparse quality gain over top-1. \\
Top-2 MoE & 40.5M tokens & yes & yes & Final val $2.1619$; similar loss, lower router log-z and milder sharpness. \\
\bottomrule
\end{tabular}%
}
\end{table}

The headline comparison is the completed three-seed full-data study across the dense active-match baseline, the stabilized top-2 MoE, and the dense total-match baseline. The ordering is stable across all three seeds. The dense active-match model reaches $1.6545 \pm 0.0012$ best validation loss, the MoE reaches $1.5788 \pm 0.0020$, and the dense total-match model reaches $1.5608 \pm 0.0025$. In perplexity terms, this corresponds to $5.231 \pm 0.006$, $4.849 \pm 0.010$, and $4.763 \pm 0.012$, respectively.

\begin{table}[!htbp]
\centering
\small
\caption{Headline three-seed full-data result, reported as mean $\pm$ standard deviation over seeds using the best validation loss from each run. The MoE clearly beats the matched-active dense baseline, while the dense total-match baseline remains slightly stronger in absolute terms.}
\label{tab:headline}
\begin{tabular}{lcccc}
\toprule
Model & Total Params & Active Params & Val Loss & PPL \\
\midrule
Dense active-match & $\approx$14.89M & $\approx$14.89M & $1.6545 \pm 0.0012$ & $5.231 \pm 0.006$ \\
MoE top-2 + z-loss & $\approx$21.06M & $\approx$14.77M & $1.5788 \pm 0.0020$ & $4.849 \pm 0.010$ \\
Dense total-match & $\approx$21.06M & $\approx$21.06M & $1.5608 \pm 0.0025$ & $4.763 \pm 0.012$ \\
\bottomrule
\end{tabular}
\end{table}

These numbers imply a matched-active gap of $0.0758 \pm 0.0021$ in the MoE's favor and a matched-total gap of $0.0180 \pm 0.0020$ in the dense model's favor. This is the core empirical picture of the paper: sparse training gives a clear gain when fairness is defined by active per-token capacity, while dense training remains slightly stronger when given the same total stored parameter budget.

The curve-level evidence sharpens that picture. Across training budgets from 40.5M to 426.3M tokens contributing to next-token loss, the matched-active advantage grows while the matched-total dense advantage narrows sharply. The sparse model consistently sits between the two dense baselines: it is better than the equal-active dense model but still slightly worse than the equal-total dense model. This pattern is exactly what one would expect if MoE's main value in this regime is conditional use of additional stored parameters rather than absolute total-capacity superiority.

\begin{figure}[!htbp]
\centering
\IfFileExists{figures/main_full_curves.pdf}{
  \includegraphics[width=\linewidth]{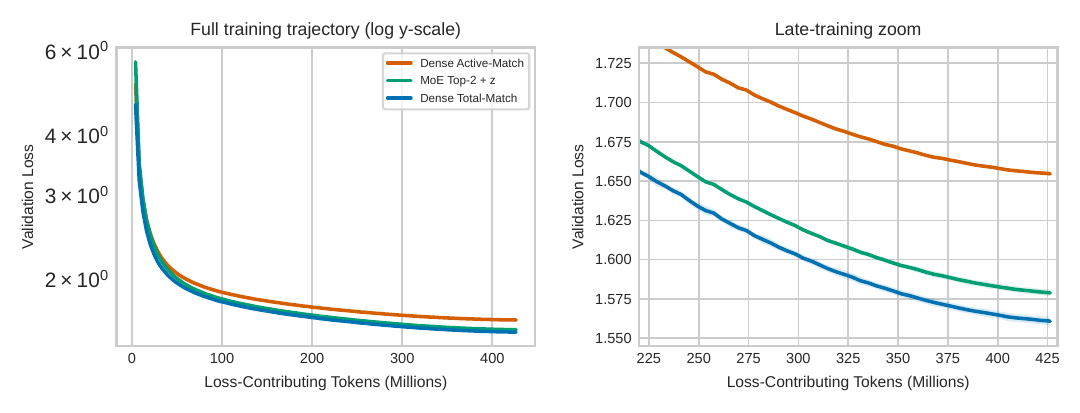}
}{
  \fbox{\parbox[c][2.0in][c]{0.95\linewidth}{\centering Placeholder for Figure 1\\validation loss vs loss-contributing tokens\\three-seed mean with $\pm 1$ std}}
}
\caption{Validation loss versus tokens contributing to next-token loss for the three headline models on full TinyStories, with each curve showing the mean over three seeds and the shaded region showing $\pm 1$ standard deviation. The MoE consistently outperforms the matched-active dense baseline, while the matched-total dense baseline remains slightly stronger in absolute validation loss.}
\label{fig:main-curves}
\end{figure}

\begin{figure}[!htbp]
\centering
\IfFileExists{figures/fairness_gap_curves.pdf}{
  \includegraphics[width=\linewidth]{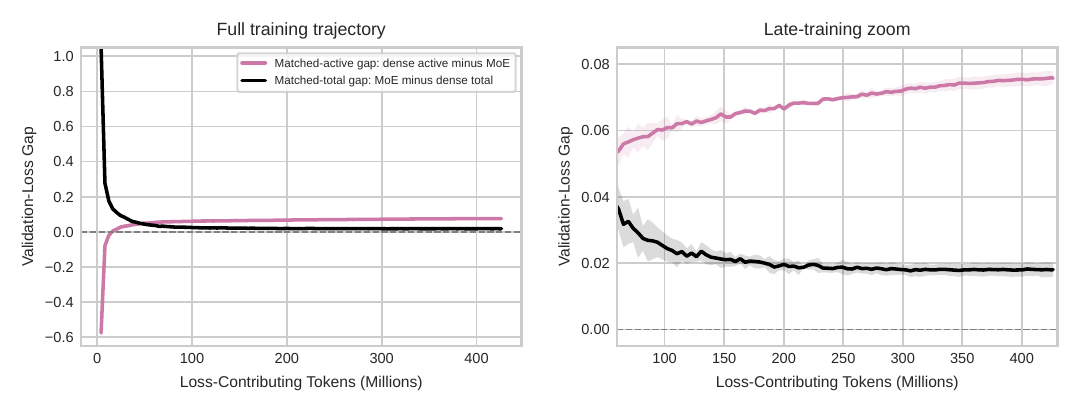}
}{
  \fbox{\parbox[c][2.0in][c]{0.95\linewidth}{\centering Placeholder for Figure 2\\fairness-gap curves over training}}
}
\caption{Validation-loss gap over training for the two fairness comparisons, again using three-seed means with shaded $\pm 1$ standard deviation. The matched-active gap (dense active minus MoE) grows in the MoE's favor with training, while the matched-total gap (MoE minus dense total) shrinks substantially but remains positive.}
\label{fig:gap-curves}
\end{figure}

Routing diagnostics support the interpretation that the sparse gain is not coming from degenerate or collapsed routing. To test whether the sparse gain arises from stable routing rather than collapse, we report compact routing diagnostics covering busiest-expert fraction, expert-usage variance, entropy by layer, and router log-z by layer. Healthy diagnostics across these views strengthen the claim that the sparse improvement reflects a real modeling gain under the matched-active comparison rather than a pathological routing artifact.

\begin{figure}[!htbp]
\centering
\IfFileExists{figures/routing_main.pdf}{
  \includegraphics[width=\linewidth]{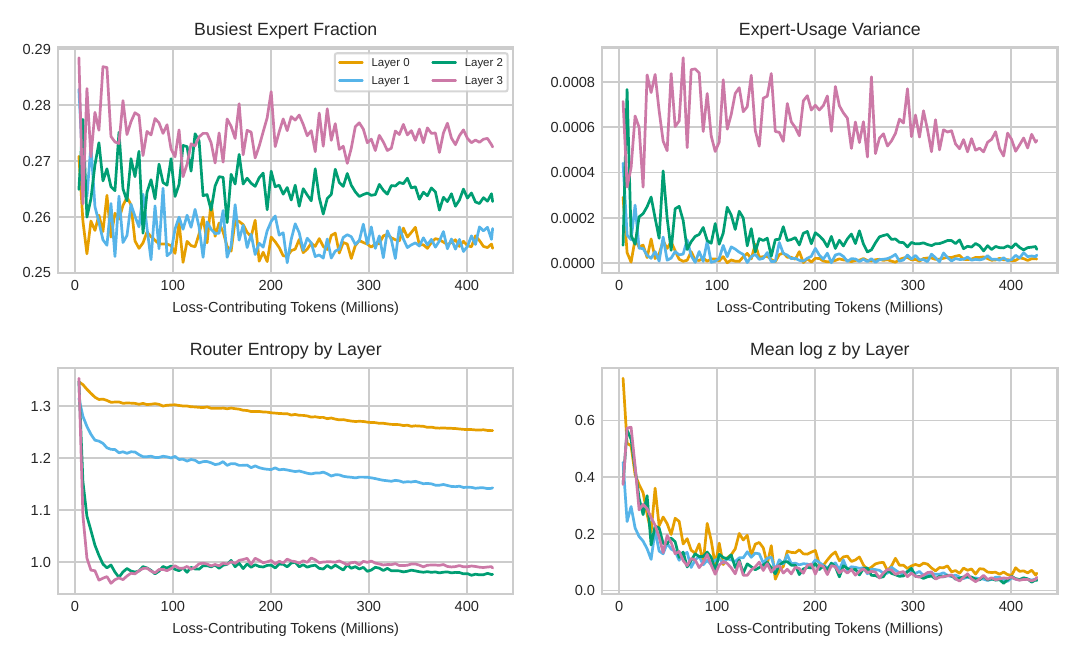}
  \vspace{-0.8\baselineskip}
}{
  \fbox{\parbox[c][2.0in][c]{0.95\linewidth}{\centering Placeholder for Figure 3\\routing stability diagnostics}}
}
\caption{Routing diagnostics for the full-data MoE run, showing busiest-expert fraction, expert-usage variance, router entropy, and mean log-z over training. Expert loads remain balanced while deeper layers become more selective, indicating that the sparse gain is achieved in a stable routing regime rather than through collapse or pathological imbalance.}
\vspace{0.2\baselineskip}
\label{fig:routing-main}
\end{figure}

Qualitative generations are anecdotal support only. We do not report judge-based sample-quality scores because calibration was not reliable enough to strengthen the paper.

\FloatBarrier

%% file: sections/limitations.tex
\section{Limitations}

This study is intentionally narrow. It focuses on TinyStories, a tiny-scale decoder-only training regime, and a budget-matched dense-versus-sparse comparison rather than an iso-width backbone swap. The sparse model changes the feed-forward sublayer, while the dense baselines are modestly width-resized to hit the active- and total-parameter budgets. We therefore do not claim that the observed matched-active advantage or matched-total near-parity automatically extends to broader corpora, larger parameter regimes, or downstream transfer settings.

We also do not claim a wall-clock sparse-training win from the main modeling experiments. Systems-oriented throughput analyses, including grouped and stacked dispatch variants, are reported in the appendix as useful context, but the central claims of the paper come from the canonical dense-versus-sparse training runs rather than from optimized sparse kernels or deployment measurements. Likewise, we do not claim that the sparse model overtakes dense training at matched total parameters; the evidence supports narrowing, not crossover.

Finally, the paper presents evidence consistent with dormant-capacity use under conditional computation, but not a causal proof of that mechanism. The main contribution is a controlled mapping of how dense-versus-sparse conclusions depend on the fairness criterion in a tiny-scale regime.

%% file: sections/conclusion.tex
\vspace{-0.7\baselineskip}
\section{Conclusion}
\enlargethispage{2\baselineskip}
\vspace{-0.3\baselineskip}

We presented a controlled tiny-scale study of dense and mixture-of-experts transformers in a LLaMA-style pretraining stack. Routed feed-forward experts produced a stable and competitive sparse model only after explicit routing regularization: naive routing collapsed, Switch-style balancing stabilized training, top-2 routing gave the main sparse quality gain, and router z-loss mainly improved router behavior.

The core result is that the fairness criterion changes the answer. At matched active parameters, sparse MoE consistently outperforms the dense baseline; at matched total parameters, dense remains slightly stronger. Across token budgets, the matched-active advantage grows and the matched-total dense advantage narrows, suggesting an active-parameter-matched conditional-capacity advantage rather than a raw total-capacity advantage.

%% file: sections/appendix.tex
\section{Appendix Overview}

This appendix collects secondary but useful supporting material: a per-seed breakdown of the full-data result, compact routing diagnostics, single-GPU throughput context for alternative sparse dispatch paths, qualitative notes, and brief reproducibility notes.

\Needspace{34\baselineskip}
\subsection{Reproducibility Details}

\begin{table}[H]
\centering
\footnotesize
\caption{Architecture details for the headline full-data runs. Dense baselines are modestly width-resized for tight active- or total-parameter matching.}
\label{tab:repro-config}
\begin{tabular}{p{0.23\linewidth}p{0.69\linewidth}}
\toprule
Item & Details \\
\midrule
Base architecture & Decoder-only LLaMA-style stack with RoPE, RMSNorm ($\epsilon=10^{-5}$), SwiGLU FFNs, GQA, tied input/output embeddings, no linear biases. \\
Common run settings & 4 layers, context length 512, dropout 0.1 for attention output and FFN paths, PyTorch SDPA attention, AMP enabled on CUDA. \\
Dense active-match & $d_{\mathrm{model}}=320$, 10 query heads, 2 KV heads, FFN hidden size 1120. \\
MoE top-2 + z-loss & $d_{\mathrm{model}}=256$, 4 query heads, 2 KV heads, 4 experts, expert hidden size 1024, top-2 token-choice routing, dropless dispatch. \\
Dense total-match & $d_{\mathrm{model}}=384$, 6 query heads, 2 KV heads, FFN hidden size 1728. \\
MoE objective & $L_{\mathrm{CE}} + 10^{-2}L_{\mathrm{bal}} + 10^{-3}L_z$; dense runs use $L_{\mathrm{CE}}$ only. \\
\bottomrule
\end{tabular}
\end{table}

\begin{table}[H]
\centering
\footnotesize
\caption{Training and data details for the headline full-data runs. The same tokenizer, data shards, optimizer, schedule, evaluation code, and seed list are used across the three model families unless noted.}
\label{tab:repro-training}
\begin{tabular}{p{0.23\linewidth}p{0.69\linewidth}}
\toprule
Item & Details \\
\midrule
Optimizer/schedule & AdamW, betas $(0.9, 0.95)$, weight decay 0.1, max LR $3{\times}10^{-4}$, min LR $3{\times}10^{-5}$, 3\% warmup then cosine decay. \\
Batching & Batch size 16, gradient accumulation 2, 26,073 optimizer steps for full-data runs, gradient clipping 1.0. \\
Data/tokenizer & TinyStories train split, SentencePiece BPE vocabulary 30,008, fixed non-overlapping 512-token windows, 95/5 train/val split with sharding seed 1337. \\
Token counts & 834,322 train windows and 43,934 validation windows; 426.3M train tokens and 22.5M validation tokens contribute to next-token loss. \\
Eval/checkpointing & Evaluation every 250 optimizer steps plus epoch end; headline values use best validation loss per run. \\
Seeds & 1337, 1338, 1339; seeds control Python, NumPy, PyTorch, data-loader shuffling, and model initialization. \\
Hardware & Single local NVIDIA RTX 3060 12GB GPU; full-data training runs took roughly 8.8--9.6 hours per model including evaluation overhead. \\
\bottomrule
\end{tabular}
\end{table}

\begin{table}[H]
\centering
\small
\caption{Parameter accounting used for the fairness comparisons. Active parameters include shared non-expert parameters plus the dense or routed top-2 FFN path exercised per token; router weights are below the displayed rounding.}
\label{tab:param-accounting}
\begin{tabular}{lrrrr}
\toprule
Model & Embedding & Non-FFN Blocks & FFN/Expert Total & Total / Active \\
\midrule
Dense active-match & 9.60M & 0.99M & 4.30M & 14.89M / 14.89M \\
MoE top-2 + z-loss & 7.68M & 0.79M & 12.58M & 21.06M / 14.77M \\
Dense total-match & 11.52M & 1.58M & 7.96M & 21.06M / 21.06M \\
\bottomrule
\end{tabular}
\end{table}

\FloatBarrier

\Needspace{22\baselineskip}
\subsection{Per-Seed Full-Data Breakdown}

\begin{table}[H]
\centering
\small
\caption{Per-seed full-data validation losses for the three headline models. The gap sizes are close across seeds, which argues against the main result being a lucky-seed artifact.}
\label{tab:per-seed}
\begin{tabular}{lccccc}
\toprule
Seed & Dense Active & MoE Top-2 + z & Dense Total & Active Gap & Total Gap \\
\midrule
1337 & 1.6554 & 1.5774 & 1.5615 & 0.0780 & 0.0159 \\
1338 & 1.6532 & 1.5779 & 1.5580 & 0.0753 & 0.0199 \\
1339 & 1.6551 & 1.5811 & 1.5629 & 0.0740 & 0.0183 \\
\bottomrule
\end{tabular}
\end{table}
\FloatBarrier

\Needspace{22\baselineskip}
\subsection{Routing Diagnostics}

\begin{table}[H]
\centering
\small
\caption{Illustrative routing diagnostics from the 150M-token run. Early layers remain more diffuse while later layers become more selective, consistent with stable specialization rather than collapse.}
\label{tab:routing-diag}
\begin{tabular}{lcccc}
\toprule
Layer & Entropy & Gate & Top1--Top2 Margin & Logz Trend \\
\midrule
L0 & 1.243 & 0.443 & 0.175 & 0.623 $\rightarrow$ 0.094 \\
L1 & 1.127 & 0.533 & 0.289 & 0.373 $\rightarrow$ 0.057 \\
L2 & 0.938 & 0.644 & 0.441 & 0.403 $\rightarrow$ 0.038 \\
L3 & 0.940 & 0.645 & 0.451 & 0.348 $\rightarrow$ 0.073 \\
\bottomrule
\end{tabular}
\end{table}

\begin{figure}[H]
\centering
\IfFileExists{figures/routing_dynamics.pdf}{
  \includegraphics[width=\linewidth]{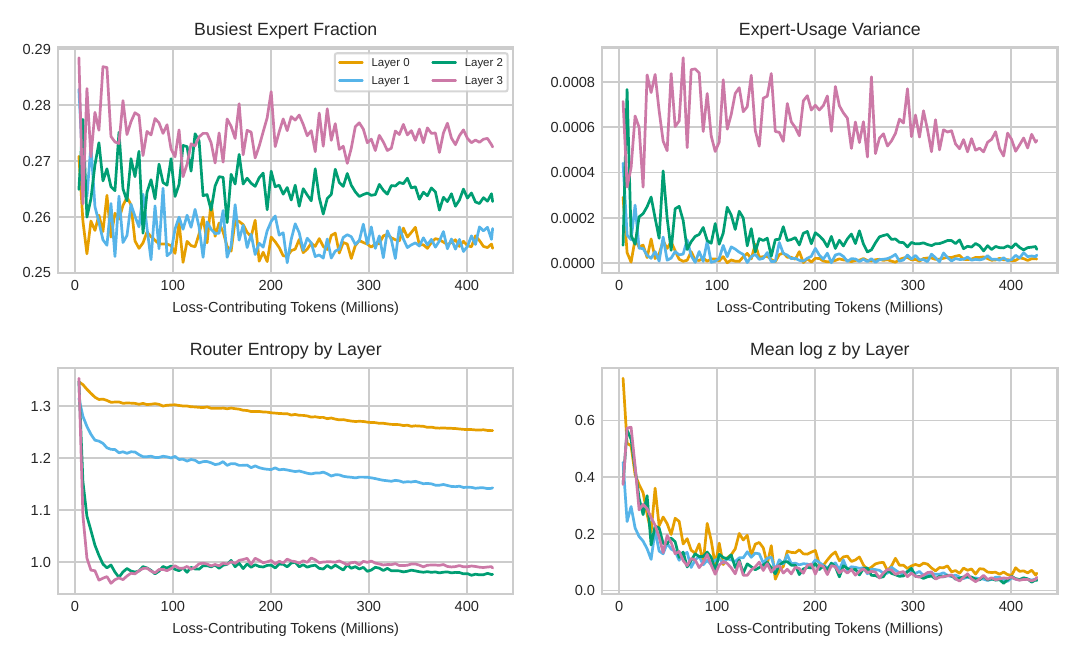}
}{
  \fbox{\parbox[c][1.9in][c]{0.95\linewidth}{\centering Placeholder for routing-dynamics figure\\busiest expert fraction, usage variance, entropy by layer, logz by layer}}
}
\caption{Routing diagnostics for the full-data MoE run. Expert loads remain balanced while deeper layers become more selective, indicating that the sparse gain is achieved in a stable routing regime rather than through collapse or pathological imbalance.}
\label{fig:routing-appendix}
\end{figure}
\FloatBarrier

\Needspace{22\baselineskip}
\subsection{Single-GPU Throughput Context}

\begin{table}[H]
\centering
\small
\caption{Single-GPU training throughput context. Optimized grouped and stacked dropless dispatch materially improve sparse throughput relative to the original naive path, but the main paper's core claim remains modeling rather than an end-to-end wall-clock win.}
\label{tab:throughput}
\begin{tabular}{lcc}
\toprule
Config & Training Throughput & Source \\
\midrule
MoE naive & 42.8k--48.4k tok/s & parity/throughput notes \\
MoE grouped & 54.4k--57.6k tok/s & trainer probes \\
MoE stacked & 61.0k--61.7k tok/s & supplementary measurement \\
Dense active-match & $58.1 \pm 0.1$k tok/s & three full seeds \\
MoE top-2 + z-loss & $54.7 \pm 0.4$k tok/s & three full seeds \\
Dense total-match & $59.3 \pm 1.2$k tok/s & three full seeds \\
\bottomrule
\end{tabular}
\end{table}

\begin{figure}[H]
\centering
\IfFileExists{figures/throughput_dispatch.pdf}{
  \includegraphics[width=\linewidth]{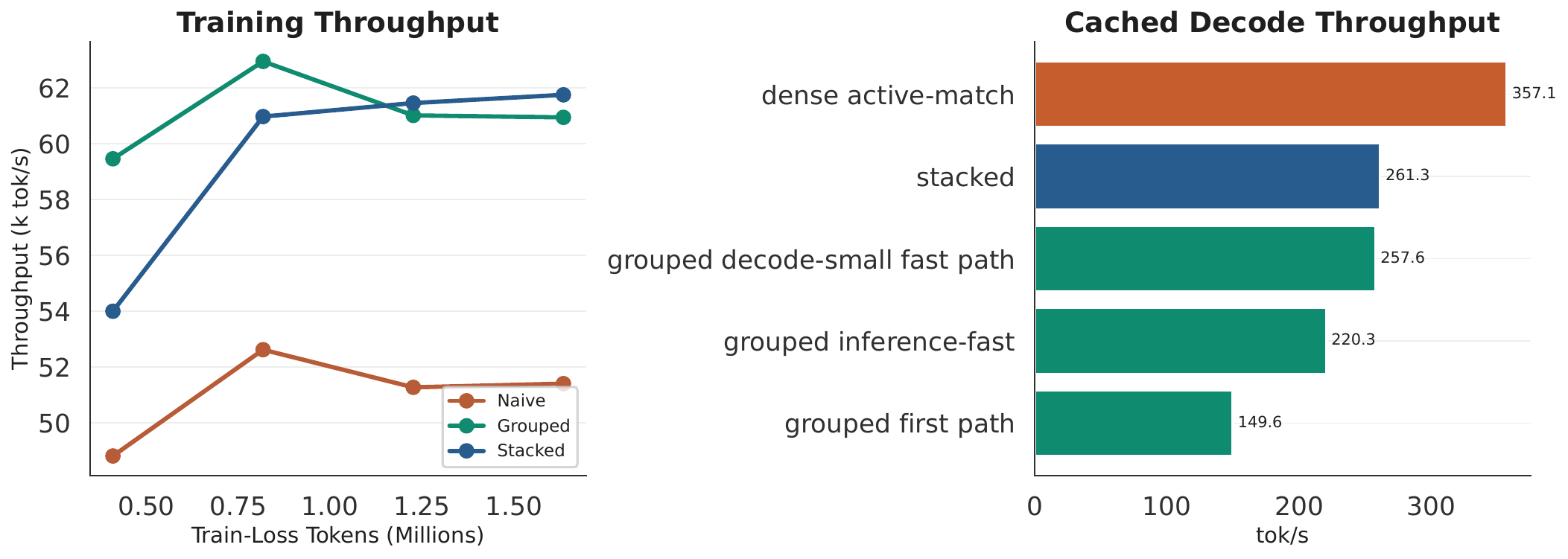}
}{
  \fbox{\parbox[c][1.9in][c]{0.95\linewidth}{\centering Placeholder for throughput appendix figure\\naive vs grouped vs stacked sparse dispatch}}
}
\caption{Single-GPU sparse throughput under naive, grouped, and stacked dispatch. Grouped and stacked dispatch materially improve sparse throughput, and these results are included as appendix-side context rather than the basis of the main modeling claim.}
\label{fig:throughput}
\end{figure}
\FloatBarrier

\subsection{Qualitative Notes}

Qualitative generations broadly line up with the metric ordering, but they remain anecdotal. The dense total-match model is usually the strongest absolute model, the dense active-match model is the weakest, and the MoE model typically falls between them. We do not report judge-based sample-quality scores because the current judge calibration was not reliable enough to strengthen the paper.
\FloatBarrier

\subsection{Reproducibility Notes}

The main seeded results come from the canonical three-seed full-data runs used throughout the paper. The corresponding configs, logs, and exported metrics were maintained for analysis and figure generation during the study. The paper's central modeling claims are based on those canonical runs, while throughput and dispatch comparisons remain secondary appendix-side context.